\documentclass[runningheads]{llncs}
\usepackage{graphicx}
\usepackage{adjustbox}
\usepackage{xspace}
\usepackage{xcolor}
\usepackage{comment}
\usepackage{cite}
\usepackage{caption} 
\usepackage[utf8]{inputenc}
\usepackage{multirow}
\usepackage{subfig}
\usepackage{tikz}
\usepackage{float}
\usepackage{arydshln}
\usepackage{array}
\usepackage{booktabs}
\usepackage{tabularx}
\usepackage{graphicx}
\usepackage{nicematrix}
\restylefloat{table}
\usepackage{latexsym}

\PassOptionsToPackage{bookmarks=false}{hyperref}
\usepackage{latexsym}
\usepackage{amsmath, amssymb}
\usepackage{url}

\usepackage[belowskip=-10pt,aboveskip=10pt]{caption}
\PassOptionsToPackage{bookmarks=false}{hyperref}

\begin{document}
\mainmatter 
\title{Abstractive Text Summarization using Attentive GRU based Encoder-Decoder}
%
\author{Tohida Rehman\inst{1}\thanks{corresponding author} \and Suchandan Das\inst{1}  \and
Debarshi Kumar Sanyal \inst{2}\and\\ Samiran Chattopadhyay\inst{3}}

\authorrunning{Rehman, et al.}
\titlerunning{Rehman, et al.}
%

\institute{Jadavpur University, Kolkata, India.\\ \email{\{tohida.rehman, suithit\}@gmail.com} 
\and
Indian Association for the Cultivation of Science, Kolkata, India.\\
\email{debarshisanyal@gmail.com} \and TCG Crest; Jadavpur University, Kolkata, India.\\
\email{samirancju@gmail.com}}

\maketitle              
\begin{abstract}
In today’s era huge volume of information exists everywhere. Therefore, it is very crucial to evaluate that information and extract useful, and often summarized, information out of it so that it may be used for relevant purposes. This extraction can be achieved through a crucial technique of artificial intelligence, namely, machine learning. Indeed automatic text summarization has emerged as an important application of machine learning in text processing. In this paper, an english text summarizer has been built with GRU-based encoder and decoder. Bahdanau attention mechanism has been added to overcome the problem of handling long sequences in the input text. A news-summary dataset has been used to train the model. The output is observed to outperform competitive models in the literature. The generated summary can be used as a newspaper headline. 

\end{abstract}

\section{Introduction}
The quantity of data around us is increasing at such a high velocity that we all need a mechanism to access correct and quick information that cuts through the noise and is brief enough to be assimilated yet not lacking in crucial content. We need a method to obtain a correct summary from an outsized volume of data. 
Automatic text summarization is such a technique through which a large chunk of information can be condensed into a meaningful summary. Extractive and abstractive summarization are two types of text summarization methods. A technique for \textit{extracting} essential sentences or paragraphs from the source text and condensing them into a shorter text is known as extractive summarization. The statistical and linguistic properties of sentences, as well as their extraction and placement in the output text, are used to determine the relevance of sentences. An abstractive summarization technique tries to present the text's primary idea in natural language without the verbatim use of terms from the text. 
The original text is transformed into a more comprehensible conceptual form in the abstractive summary approach, resulting in a shorter summary of the original text content.


In this paper, we present an encoder-decoder based model to summarize documents. A gated recurrent unit (GRU) has been used to boost a recurrent neural network's memory capacity as well as to make training a model easier. It also helps us to overcome the vanishing gradient problem. In attention mechanism, the context vector concatenated with the previous decoder output. That are fed along with previous decoder hidden state into the Decoder GRU component for each time step to generate the output \cite{bahdanau2014neural}.
We have used the CNN/Daily Mail dataset\cite{nallapati2016abstractive,see2017get}. We obtained higher F1 scores using ROUGH-1 and ROUGH-L compared to some other competitive baselines in the literature.

\section{Related Works}
Nallapati et al \cite{nallapati2016abstractive} has proposed baseline encoder and decoder architecture where LSTM has been used. Bidirectional as well as unidirectional LSTM was used at encoder and decoder correspondingly. Word level and sentence level bidirectional GRU was used. Performance of basic encoder and decoder model  has been improved through Bahdanau et al \cite{bahdanau2014neural}. See et al. \cite{see2017get} offered a detailed study of numerous abstractive text summarization models for pointer-generator and RNN seq2seq models that are based on sequence-to-sequence encoder-decoder architecture. Sutskever et al. \cite{sutskever2014sequence} proposed a multilayer LSTM based end-to-end solution to sequence learning. The input for the encoder was a fixed length of text, and the output for the decoder was the same. Lin et al\cite{lin2018global} proposed global encoding mechanism of abstractive text summarization. In this paper, we have designed GRU based encoder and decoder with one extra attention layer. Shi et al \cite{shi2021neural} proposed to ``improve seq2seq models, making them capable of handling different challenges, such as saliency, fluency and human readability, and generate high-quality summaries". Generally speaking, most of these techniques differ in one of these three categories: network structure, parameter inference, and decoding/generation. Luong et al \cite{luong2015effective} examines two simple and effective classes of attention mechanism: a global approach which always attends to all source words and a local one that only looks at a subset of source words at a time. Ksenov et al \cite{aksenov2020abstractive} proposed ``the encoder and decoder of a Transformer-based neural model on the BERT language model". Recently, a model proposed as ``BART: Denoising Sequence-to-Sequence Pre-training for Natural Language Generation, Translation, and Comprehension" \cite{lewis2019bart} which captures the simplicity of BERT (Devlin et al.)\cite{devlin2018bert} and GPT (Radford et al.) \cite{radford2018improving} and others pre-training schemes. BART opens many ways to thinking for fine-tuning in text summarization application.

\section{Methodology}
\begin{figure}[H]
\centering 
\includegraphics[scale=0.60]{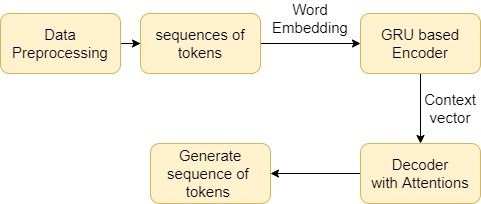}
\caption{Generic work flow of our model.}
\centering
\label{fig:steps}
\end{figure}

In this section, we describe the methodology that we have used to design our abstractive text summarizer.
Generic work flow of our model shown in Fig.\ref{fig:steps}. Here we used GRU \cite{DBLP:journals/corr/ChoMBB14} in Seq2seq model.
The GRU has gating units to manage flow of information inside the unit. 

Several crucial steps were followed such as data collection and pre-processing, tokenization, encoder and decoder model design, training the model, evaluation of the model and so on to overcome text generation problem to predict proper semantics meaningful summary.

\noindent Let consider input sequence is like below
\begin{equation}
I=X_1, X_2 ----------X_d
\end{equation}
Where, d is the vocabulary size of input sequence for above mentioned input sequence, the output sequence will be like
\begin{equation}
O=Y_1, Y_2 ----------Y_s
\end{equation}
Where s is the vocabulary size of output sequence. Here, s$<$d, it means length of output sequence is less than the length of input sequence.

\subsection{Data collection and pre-processing}
\begin{figure}[H]
\centering 
\includegraphics[scale=0.67]{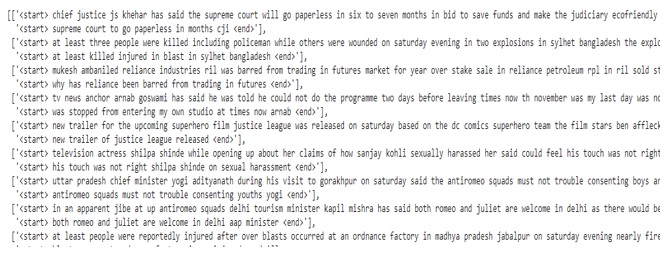}
\caption{Cleaned news summary data.}
\centering
\label{fig:cleaneddata}
\end{figure}

Dataset plays a key role in each and every deep learning process. To get better result, it is very important to get good dataset. Various type of data sets is present in different resources. We have used the CNN/Daily Mail dataset\cite{nallapati2016abstractive,see2017get}. There are different columns present in the data set but we have taken news and summary description to fulfill our purpose. Due to low configuration of our system, 10000 examples from CNN/Daily Mail dataset has been used.\\
Before we begin creating the model, we must first complete some basic pre-processing tasks. A decision based on messy and filthy text understanding could be disastrous. As a result, we have removed all unneeded symbols, letters, and other elements from the text that do not affect the target of our downside throughout this phase. We have removed HTML tags, parenthesis, and special character.
\begin{figure}[H]
\centering 
\includegraphics[scale=0.60]{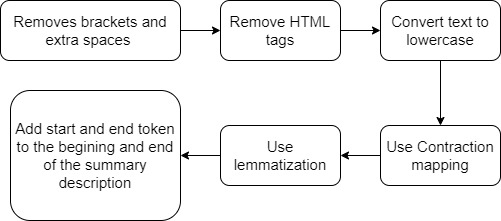}
\caption{Steps used in data pre-processing.}
\centering
\label{fig:preprocessing}
\end{figure}
To begin with, we changed the entirety of the content to lower case, and afterward we split it up \cite{masum2019abstractive}. There is different constriction in the English language, for example, doesn't, aren't, etc. We have added contractions mapping in pre-processing phase. We have removed unnecessary components from the raw text to get the cleaned text. Then, at that point we lemmatized the words that have various types of a similar term. At the beginning and end of the news and summary description, we have included START and END tokens. 
Fig. \ref{fig:preprocessing} represents steps that we used to clean data set to prepare as news abstract and summary pair. 
Fig. \ref{fig:cleaneddata} refer some cleaned data.\\

\subsection{GRU based encoder-decoder with attention}
Cho et al \cite{cho2014learning} introduced the RNN based encoder-decoder where RNN in encoder helps to encodes sequence of words into a fixed length vector representation and in other hand RNN in decoder helps to decode the incoming representation into a sequence of words.
We used a bidirectional GRU
encoder, a unidirectional GRU decoder with attention mechanism \cite{nallapati2016abstractive}.
Here, seq2seq model \textit{with attention mechanism}\cite{bahdanau2014neural} builds a context vector using all hidden states present in the encoder.
It aids in focusing on the most important information in the source sequence. The decoder uses the context vectors associated with the source position and the previously created target words to predict the target word at each time stamp.
Below are the steps which describe how the Bahdanau attention mechanism works\cite{bahdanau2014neural}.\\
\begin{enumerate}
\item The encoder produces the annotation
 $(h_{i})$ for each word $x_{i}$, for an input sentence of length T words at each time step i. Encoder has bidirectional GRU, reads the input sentence in forward as well as in backward direction to generate the $(h_{i})$ for each time steps. 
 \begin{equation}
 h_{i}= [\overrightarrow{h_{i}^T},\overleftarrow{h_{i}^T}]^T
\end{equation}

\item  At each time step, the decoder takes the annotations $(h_{i})$ and the previous hidden states $s_{i-1}$ to calculate attention score $(e_{ij})$.  It can be written as follows.
\begin{equation}
     e_{ij} = att (s_{i-1}, h_j) 
\end{equation}
Bahdanau et al. is referred to as additive attention is defined below: 
\begin{equation}
     att (s_{i-1}, h_j) = V^\top \tanh (W [s_{i-1}, h_j])  
\end{equation}
Where $W$, $V$ are the trainable weights.
\item The attention weights $(\alpha_{ij})$ are computed as follows:        
\begin{equation}
\alpha_{ij}=\frac{exp(e_{ij})}{\sum_{k=1}^{T_x} exp(e_{ik})}
\end{equation}
\item Linear sum is computed using attention weight $(\alpha_{ij})$ and hidden state of encoder to generate the context vector. This context vector is calculated as follows:
\begin{equation}
    c_i=\sum_{j=1}^{T_x}{\alpha_{ij}}h_j
\end{equation}
\item At time step $i$, the decoder produces the hidden state $(s_i)$ depending upon  $s_{i-1}$ which is the previous hidden state, $y_{i-1}$ which is the target word at time step $i-1$, and $c_i$ which is the context vector. 
\begin{equation}
s_{i}=f(s_{i-1},y_{i-1},c_{i})
\end{equation}

\item Steps 2 to 5 are repeated until the end of the sentence or the maximum length of generated tokens is reached.
Each word is predicted based on the following rule: 
\begin{equation}
P(y_i|y_{i-1},y_{i-2}....y_1,X)=g(y_{i-1},s_i,c_i)
\end{equation}
\end{enumerate}

                              				                  
 
 
\begin{figure}[H]
\centering 
\includegraphics[scale=0.98]{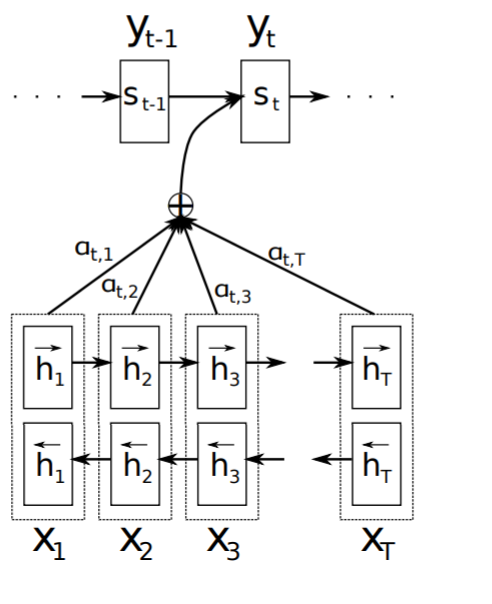}
\caption{How attention works in seq2seq Encoder Decoder model. Adopted from \cite{bahdanau2014neural}}
\centering
\label{fig:attention}
\end{figure}

In Fig \ref{fig:attention}, it shows how attention works in sequence to sequence encoder-decoder model based on GRU.
\section{Experiment and Result analysis}
As the computational power of our machines was low, a small dataset has been used. Here, we have used 10000 examples from CNN/Daily Mail dataset\cite{nallapati2016abstractive,see2017get}, Adam optimizer, a Sparse Categorical Cross entropy loss function with batch size =128, embedding dimension = 256, hidden units = 1024. 
We used $80\%$ of the data for training purposes and $20\%$ for testing purposes.
We have trained the model for 100 epochs. Loss has been reduced to 0.0480.  Table \ref{Table:table1} shows F1 of ROUGH-1 and ROUGH-L score on the basis of the output from the model. We now provide some illustrative examples of the output of our model.\\

\begin{table*}[!htbp]
\begin{center}
{ \begin{tabular}{|c|c|}\hline
\texttt{ROUGH-1} & \texttt{ROUGH-L} \\ \hline
F1 & F1 \\ \hline
\texttt{35.29} & \texttt{35.25} \\\hline
\end{tabular} }
\vspace*{2mm}
\caption{ROUGH score (F1) on the basis of output from model.}
\label{Table:table1}
\end{center}
\end{table*}
\vspace{-1cm}
\subsection{Sample Output}
\paragraph{\textbf{Input}: ``actress deepika padukone has said that she will not be walking the red carpet at the cannes film festival deepika added right now all my energies are focused on padmavati earlier it was reported that deepika had been ap-pointed the brand ambassador of oral and would represent the brand at the film festival”}

\paragraph{\textbf{Actual Summary}: deepika padukone will not be walking the red carpet at the cannes film festival.}

\paragraph{\textbf{Predicted Summary}: not walking red carpet at cannes film festival says deepika.}

\paragraph{\textbf{Input}: “beverage giant pepsico ceo indra nooyi received million over crore in compensation for marking increase in her pay this was the fourth consecutive pay raise for nooyi who has been the ceo since the rise in compensation came as efforts to steer the companys port-folio away from sugary products helped earnings”}

\paragraph{\textbf{Actual Summary}: pepsico ceo indra nooyi received million over crore.}

\paragraph{\textbf{Predicted Summary}: pepsico ceo indra nooyi pay rises to crore in year.}

\subsubsection{Heatmap:}
\begin{figure}[H]
\centering 
\includegraphics[scale=0.4]{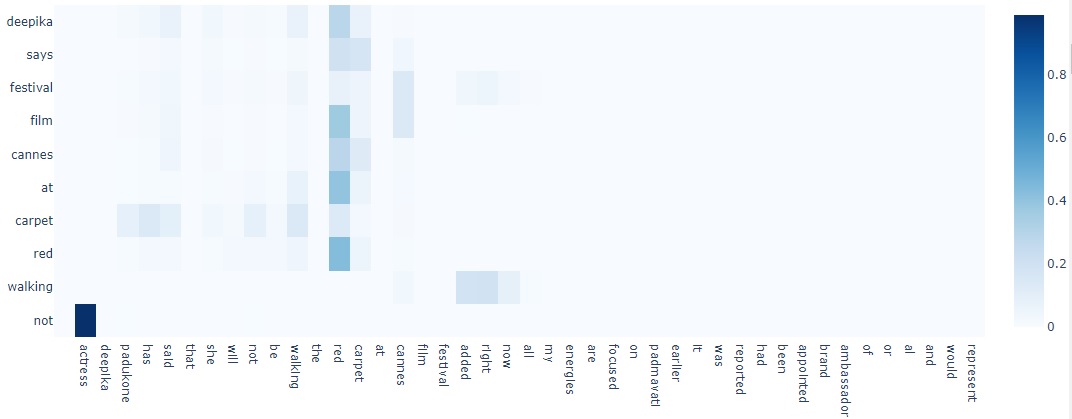}
\caption{Attention Heatmap.}
\centering
\label{fig:attentionHeat map}
\end{figure}

Heatmaps for predictive outputs are given below figures which are more interesting. In attention heatmap plot, x axis denotes the actual input, y axis denotes the summary output and z axis indicates attention plot weight. Main goal of using attention mechanism is to emphasize on the important information. In Fig. \ref{fig:attentionHeat map} shows which parts of the input sentence has the model's attention while generating summary.

In our proposed solution,  we have used  daily news dataset get  35.29 ROUGH 1 score and 35.25 ROUGH L score as F1 score which is slightly better than some existing model. It generates more semantics meaningful single sentence summary. We have also tried to compare the model performance with other existing model using ROUGH score. Below Table \ref{Table:table2} shows the comparison of ROUGH 1 and ROUGH L scores with some existing model. k refers to the size of the beam for generation.

\begin{table*}
\centering
\begin{tabular}{|c|c|c|}\hline
\multirow{2}{4em}{Model} & {ROUGH-1} &  {ROUGH-L}\\ \cline{2-3}
     & {F1} &  {F1}\\ \hline
Words-lvt5k-1sent \cite{nallapati2016abstractive}&28.61& 	25.423\\
\hline
Words-lvt2k-temp-att \cite{nallapati2016abstractive}&35.46 &	32.65\\ \hline
ABS+ (Rush et al.)\cite{DBLP:journals/corr/RushCW15}&	28.18 &	23.81\\\hline
RAS-Elman (k=10)(Chopra et al.) \cite{chopra2016abstractive}&33.78 &	31.15\\\hline
Our Model & \textbf{35.29} &	\textbf{35.25}\\ \hline
\end{tabular} 

\vspace*{2mm}
\caption{Comparison of the ROUGH score (F1) with some existing model}
\label{Table:table2}
\end{table*}
\vspace{-1 cm}
\section{Conclusion and Future Work}
GRU-based encoder and decoder model with Bahdanau attention mechanism has been used to design an automatic text summarizer. The attention mechanism also emphasizes the important word of the sequence and copy the same in the output summary. The proposed method provides better result than several other approaches in the literature. 
A meaningful summary with single sentence has been generated which can be used for news headline generation. However, we also observed that our model is not always producing the best result. In future, we will use BERT based pre-training model to enhance model's performance and to generate more meaningful summary. We will try to create summary of Covid-19 related scientific articles which can help the medical community by providing a clean and meaningful high-quality knowledge base of the pandemic. 
\bibliographystyle{IEEEtran}
\bibliography{ref}
\end{document}